\documentclass[letterpaper, 10 pt, conference]{ieeeconf}  % Comment this line out if you need a4paper

\IEEEoverridecommandlockouts                              % This command is only needed if 
\usepackage{graphics} % for pdf, bitmapped graphics files
\usepackage{epsfig} % for postscript graphics files
\usepackage{mathptmx} % assumes new font selection scheme installed
\usepackage{times} % assumes new font selection scheme installed
\usepackage{amsmath} % assumes amsmath package installed
\usepackage{amssymb}  % assumes amsmath package installed\
\usepackage{url}
\usepackage{subcaption}
\usepackage{hyperref}
\usepackage{multirow}
\usepackage{caption}

\title{\LARGE \bf NV-LIO: LiDAR-Inertial Odometry using Normal Vectors \\ Towards Robust SLAM in Multifloor Environments}

\author{Dongha Chung and Jinwhan Kim$^{*}$% <-this % stops a space
\thanks{*This work was not supported by any organization}% <-this % stops a space
\thanks{$^{*}$Corresponding author: Jinwhan Kim}
\thanks{Dongha Chung and Jinwhan Kim are with the Department of Mechanical Engineering, KAIST (Korea Advanced Institute of Science and Technology), 291 Daehak-ro, Yuseong-gu, Daejeon 34141, Republic of Korea
        {\tt\small \{chungdongha, jinwhan\}@kaist.ac.kr}}%
}

\begin{document}

\twocolumn[%
\begin{@twocolumnfalse}
    % \begin{center}
        \large This work has been submitted to the IEEE for possible publication.

Copyright may be transferred without notice, after which this version may no longer be accessible.
    % \end{center}
\end{@twocolumnfalse}
]

\newpage

\maketitle
\thispagestyle{empty}
\pagestyle{empty}

%%%%%%%%%%%%%%%%%%%%%%%%%%%%%%%%%%%%%%%%%%%%%%%%%%%%%%%%%%%%%%%%%%%%%%%%%%%%%%%%
\begin{abstract}

% Indoor environments are characterized by confined spaces, thin walls delineating numerous segmented areas. Due to these environmental factors, the existing algorithms frequently encounter failures in point cloud registration, often attributed to incorrect correspondences.

Over the last few decades, numerous LiDAR-inertial odometry (LIO) algorithms have been developed, demonstrating satisfactory performance across diverse environments. Most of these algorithms have predominantly been validated in open outdoor environments, however they often encounter challenges in confined indoor settings. In such indoor environments, reliable point cloud registration becomes problematic due to the rapid changes in LiDAR scans and repetitive structural features like walls and stairs, particularly in multifloor buildings. In this paper, we present NV-LIO, a normal vector based LIO framework, designed for simultaneous localization and mapping (SLAM) in indoor environments with multifloor structures. Our approach extracts the normal vectors from the LiDAR scans and utilizes them for correspondence search to enhance the point cloud registration performance. To ensure robust registration, the distribution of the normal vector directions is analyzed, and situations of degeneracy are examined to adjust the matching uncertainty. Additionally, a viewpoint based loop closure module is implemented to avoid wrong correspondences that are blocked by the walls. The propsed method is validated through public datasets and our own dataset. To contribute to the community, the code will be made public on \url{https://github.com/dhchung/nv_lio}.

\end{abstract}

%%%%%%%%%%%%%%%%%%%%%%%%%%%%%%%%%%%%%%%%%%%%%%%%%%%%%%%%%%%%%%%%%%%%%%%%%%%%%%%%
\section{INTRODUCTION}

\begin{figure}[t]
    \centering
    \includegraphics[width=1.0\linewidth]{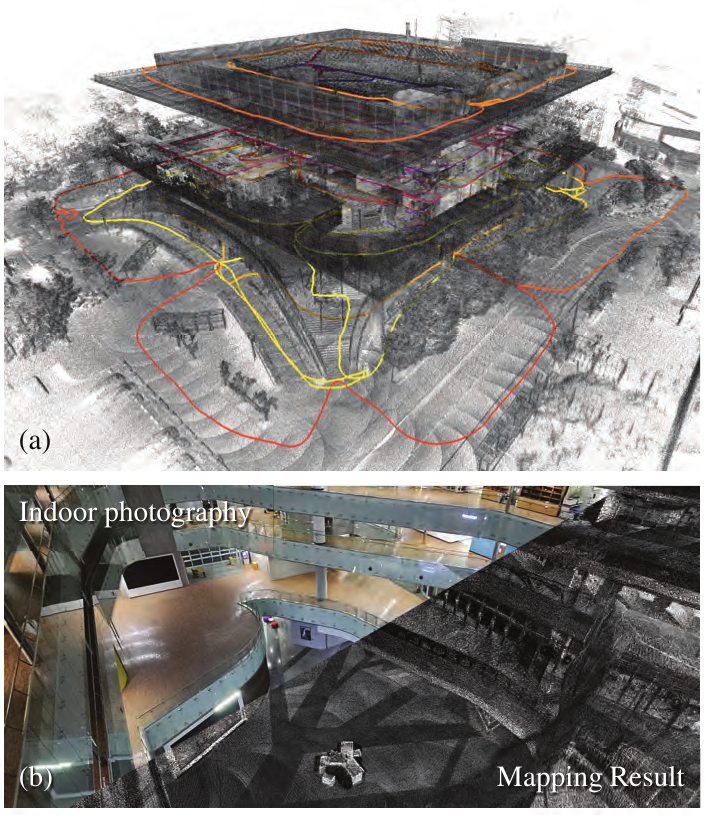}
    \caption{Mapping result of the KI building, a five-story building at KAIST, using NV-LIO. (a) shows the overall mapping result and LiDAR trajectory. (b) shows the comparison of photography taken indoors of the building with the mapping result.}
    \label{fig:ki_building}
\end{figure}

The field of mobile robotics has widened its operational environments through the assistance of advances in sensors, actuators, and control algorithms. These advancements are reshaping the robotics landscape, extending their application from autonomous driving to a diverse array of environments, encompassing firefighting robots, security robots, delivery robots, and others that are required to overcome various obstacles in confined indoor environments. Reliable SLAM techniques are required for these robots to navigate through unknown indoor environments.

LiDAR sensors directly measure the depth information of the surrounding environment, providing stable measurements regardless of the ambient brightness. Leveraging these capabilities, the 3-dimensional (3D) LiDARs are predominantly utilized for localization and mapping by registering consecutive LiDAR point clouds for mobile robotics applications. Over the last decades, numerous public datasets for autonomous driving applications have been released, which led to significant advancement of LiDAR-inertial odometry algorithm technology through the development and validation of algorithms utilizing these datasets. 

However, these algorithms are known to show degraded performance in indoor spaces and often fail significantly in multifloor environments. Unlike outdoor environments, indoor environments are characterized by confined spaces and thin walls forming multiple segmented areas. In these areas, the scene captured by the LiDAR scans may change rapidly with repetitive structural elements such as walls and stairs. Due to these environmental factors, the existing algorithms frequently encounter failures in point cloud registration. One of the reasons for such registration failures is the challenge of aligning point clouds on opposite sides of a wall, known as double-side issue \cite{PlaneSLAM}. Despite the thickness of the wall, these misalignments can inaccurately represent a wall with no thickness in the map. Another issue is the fixed parameter problems. In narrow spaces, LiDAR scans often produce a densely packed point cloud in proximity. However, when downsampling is performed using fixed parameters for matching, the number of points available for matching decreases, leading to potential inaccuracies or mismatches in the alignment process \cite{AdaLIO, AdaptiveKeyframe}.

\begin{figure*}[t]
    \centering
    \includegraphics[width=1.0\linewidth]{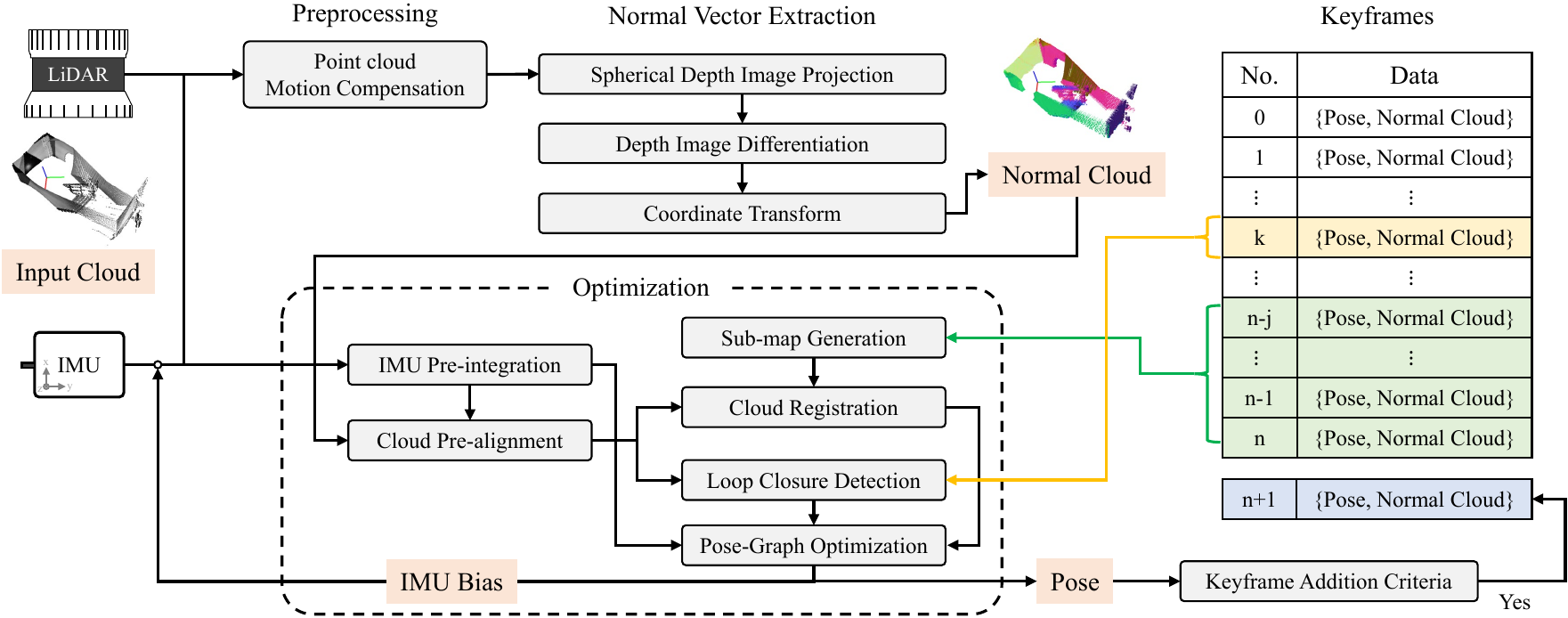}
    \caption{The framework of the proposed algorithm}
    \label{fig:algorithm_overview}
\end{figure*}

In this paper, we present NV-LIO, a normal vector-based tightly-coupled LIO framework designed for multifloor indoor SLAM. This framework utilizes the capability to project dense spinning type 3D LiDAR scans into range images, from which we extract normal vectors. The registration between scans not only considers the nearest neighbor but also takes into account the angle difference of the normal vectors, enhancing the accuracy of correspondence search during registration. During the matching process, the distribution of the normal vector directions is analyzed to assess the occurrence of degeneracy, adjusting the matching uncertainty. To ensure accurate loop closure, visibility analysis is employed during the matching process between scans and submaps, preventing incorrect correspondences between different rooms or floors. 

The main contributions of this paper can be summarized as follows:

\begin{itemize}
    \item We introduce a tightly-coupled LIO framework that utilizes normal vectors from dense mechanical LiDAR point clouds to achieve robust registration in narrow indoor environments. 
    \item To address degeneracy situations in point cloud registration, especially in long corridors or stairwell scenarios, degeneracy detection algorithm and corresponding registration uncertainty covariance matrix calculation method is proposed.
    \item We thoroughly validated the proposed algorithm in various datasets, including assessments on public datasets and our own collected dataset, as shown in Fig.~\ref{fig:ki_building}. The validation results demonstrate the effectiveness and versatility of the proposed approach.
\end{itemize}

\section{RELATED WORK}
\label{sec:related_work}

The key component of LiDAR (inertial) odometry relies on the effectiveness of point cloud registration. To handle the computational load of processing tens of thousands of points with each scan, various methods have been developed to reduce computation time. Feature-based methods extract key features from point clouds to perform matching, while direct methods involve downsampling the point cloud and using techniques such as iterative closest point (ICP) \cite{ICP} or generalized-ICP (GICP) \cite{GICP}.

Among feature-based algorithms, LOAM \cite{LOAM} utilizes 2D lidar scans with nodding motion, extracting corner points and surface points based on the relationships between adjacent points. LeGO-LOAM \cite{LEGO_LOAM} adopts LOAM's feature extraction technique for 3D LiDAR, extracting features from each ring and incorporating ground segmentation to enhance accuracy. LIO-SAM \cite{LIO_SAM} employs a similar feature extraction method but introduces tightly-coupled LIO with IMU pre-integration \cite{IMU_Preintegration}, demonstrating high performance. Fast-LIO \cite{FAST_LIO} employs the same feature extraction method but introduces a new Kalman gain formula for faster computation. 

Leveraging its speed advantage, Fast-LIO2 \cite{FAST_LIO2} introduces an direct matching method using ikd-tree \cite{ikd_tree} for fast nearest neighbor search, enabling efficient scan-to-map matching instead of traditional scan-to-scan matching, resulting in more globally consistent performance. Faster-LIO \cite{Faster_LIO}, on the other hand, replaces the tree structure of Fast-LIO2 with an incremental voxelization (iVox) system for even faster computational speeds.

While these direct methods exhibit excellent performance in outdoor environments, it is observed that they often fails in narrow indoor environments where the point clouds are densely packed in proximity. The fast computations of these algorithms rely on the downsampling of the point clouds, which may fail to preserve meaningful points for reliable registration in a cramped indoor environments. To tackle this issue, adaptive methods that adjust the parameters based on the analysis of LiDAR scans have been proposed. Based on Faster-LIO, AdaLIO \cite{AdaLIO} adaptively changes the voxelization size for downsampling, search radius, and residual margin for plane selection by analyzing the given LiDAR scan for stable registration performance in degenerate cases. In DLIO \cite{DLIO} and the work presented in \cite{AdaptiveKeyframe}, adaptive keyframe insertion methods are proposed to balance efficient computational time and reliable cloud registration. These methods insert keyframes based on scan analysis, resulting in dense keyframes in narrow spaces and sparse keyframes in wide areas.

SuMa \cite{SuMa} employs a two-step process where the LiDAR point cloud is first projected into a depth image, and normals are then extracted to form surfels. SLAM is conducted through matching between the surfel map and measured surfels, with surfel updates occurring based on the matching results. To expedite correspondence search, a rendering technique is utilized, projecting the surfel map onto the current frame to find correspondences in the image. LIPS \cite{LIPS} leverages the prevalence of planes in indoor scenarios, such as offices. It extracts planar primitives from LiDAR scans and utilizes a technique that matches these primitives. In \cite{PlaneSLAM}, the approach begins by extracting the normal vector for each point and using clustering to identify planes. A method called forward ICP flow is introduced, utilizing the point-to-plane distance to find new scan points corresponding to the existing planes, instead of finding planes in ever scan. During matching, if the angle difference between the normal vectors of the plane and the existing plane exceeds a certain threshold, the matching is avoided, effectively resolving the double-sided issue.

\section{METHOD}
\label{sec:method}

\subsection{System Overview}
The state of the system $\mathbf{X}$ is expressed as follows:
\begin{equation}
    \label{eq:state}
    \mathbf{X} = \left[R, \mathbf{t}, \mathbf{v}, \mathbf{b}_a, \mathbf{b}_g\right]
\end{equation}

where $R$, $\mathbf{t}$, $\mathbf{v}$ are the rotation matrix, position, and velocity in the global reference frame and $\mathbf{b}_a$ and $\mathbf{b}_g$ are the IMU acceleration and gyroscope bias in IMU frame. When setting a body frame different from the IMU's coordinate system, transformations of IMU acceleration and angular velocity need to be performed according to the coordinate transformation. For convenience and to reduce additional computations, the body frame of the system is aligned with the IMU frame.

The keyframe $\mathbb{K}$ is expressed as follows:
\begin{equation}
    \label{eq:keyframe}
    \mathbb{K} = \left\{\mathbf{x}(R, \mathbf{t}), N(P_{0}(\mathbf{p}_0, \mathbf{n}_0), \cdots, P_{k-1}(\mathbf{p}_{k-1}, \mathbf{n}_{k-1})) \right\}
\end{equation}

where $\mathbf{x}$ is the body pose in global reference frame when the scan is made and $N$ is the normal cloud containing normal points $P_{i}$ that consist of the point coordinates $\mathbf{p}_i$ and the normal vector $\mathbf{n}_i$. 

Unlike in outdoor environments, where a significant portion of the map is often within the LiDAR's range, the visibility of map points in the current scan may be extremely limited in narrow indoor spaces. Due to this characteristic, direct methods for matching scans directly to the map may result in drift, particularly in narrow corridors or during inter-floor transitions, making correction difficult upon returning to the same location. Therefore, in this study, we adopt a keyframe-based pose-graph SLAM framework. The map is formed as a collection of keyframes, allowing it to dynamically adjust in response to corrections.

Figure~\ref{fig:algorithm_overview} illustrates the overview of the proposed algorithm. The in-scan motion is compensated by using the attitude calculated by integrating the IMU gyroscope measurements. The normal cloud is extracted by projecting the motion-compensated cloud using spherical projection. After aligning the extracted normal cloud using inertial measurements, we determine the relative pose through normal cloud registration between the submaps composed of preceding keyframes. Additionally, correction measurements are obtained through viewpoint based loop-closure. These registration results are included in the graph as relative pose factors, and IMU measurements are also added to the graph via IMU preintegration. Through this pose-graph optimization, the current pose and IMU bias are estimated.

\subsection{Preprocessing}
To achieve accurate normal vectors and registration results, motion compensation for LiDAR scans is essential. For this purpose, we utilize the angular rate obtained by the IMU with estimated bias. Considering the higher frequency at which points from LiDAR are received (more than 10000 Hz) compared to the IMU rate (100 Hz), we perform interpolation over time using the rotation estimated by the IMU, based on the timestamp of the initial incoming point. This process is carried out in the LiDAR coordinate system to facilitate the subsequent step of spherical projection.

\subsection{Spherical Projection \& Normal Extraction}
A spherical projection is performed on the motion-compensated LiDAR point cloud to generate a depth image. In this process, the size of the depth image is manually selected, taking into account the characteristics of the LiDAR point cloud, such as the number of LiDAR channels, horizontal resolution, and field of view (FoV). With given parameters of the maximum vertical FoV ($fov_{max}$), minimum vertical FoV ($fov_{min}$), depth image height ($h$) and width ($w$), the vertical resolution is as $ver_{res} = (fov_{max}-fov_{min})/h$, and the horizontal resolution is as $hor_{res} = 2\pi/w$. The image coordinates ($u, v$) of the each point ($\mathbf{p}(x, y, z)$) is as follows:

\begin{equation} \label{eq1}
\begin{split}
u & = (\pi - atan2(y, x))/hor_{res} \\
v & = (fov_{max} - atan2(z, \sqrt{x^2 + y^2})/(ver_{res}).
\end{split}
\end{equation} 

The normal vector can be calculated by differentiating the depth value $(r(u,v))$ in horizontal direction $(\Delta r/\Delta\psi)$ and vertical direction $(\Delta r/\Delta\theta)$ of the range image as:

\begin{equation} \label{eq:normal_sp}
\mathbf{n}^s(u, v) = \begin{bmatrix}
    (\Delta r/\Delta\psi)/c \\
    -(\Delta r/\Delta\theta)/c \\
    1/c
\end{bmatrix}
\end{equation} 

where $\theta = \pi/2 - (fov_{max} - v \times ver_{res})$ represents the polar angle, $\psi = \pi - u \times hor_{res}$ represents the azimuth angle, and $c$ is a scaling variable making the normal vector as an unit vector. 

\begin{figure}[t]
    \centering
    \includegraphics[width=1.0\linewidth]{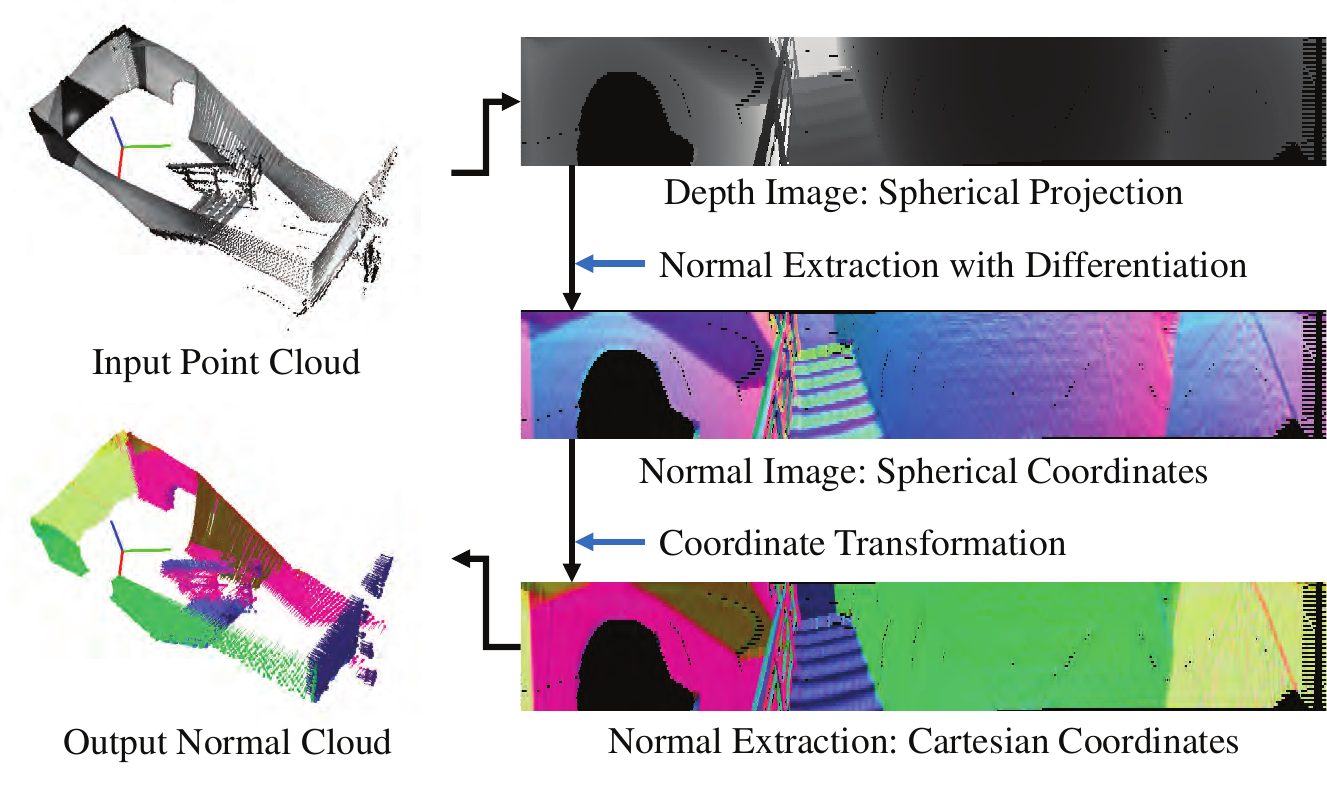}
    \caption{Normal extraction process}
    \label{fig:normal_extraction}
\end{figure}

In expansive outdoor settings, adjacent pixels cover a broader area, mitigating the influence of distance measurement noise on normal vector calculation. Conversely, in confined indoor environments, the same area is considerably smaller, amplifying the impact of distance measurement noise on the calculation outcome. Taking these considerations into account, instead of simply using differentiation between neighboring pixels, we apply a window based approach, assuming that the derivative values within the window are similar. The derivative values for pairs in each horizontal and vertical direction within the window are calculated and averaged to mitigate the effects of distance measurement noise. The normal vector represented in spherical coordinates, $\mathbf{n}^s$, is transformed into Cartesian coordinates, $\mathbf{n}^c$, based on the azimuth and elevation for each pixel, as $\mathbf{n}^c(u, v) = T(\theta,\psi) \mathbf{n}^s(u, v)$ where $T(\theta, \psi)$ is the transformation matrix as follows:

\begin{equation} \label{eq2}
T(\theta, \psi) = \begin{bmatrix}
    -\sin(\psi)&\cos(\psi)\cos(\theta)&\cos(\psi)\sin(\theta) \\
    \cos(\psi)&\sin(\psi)\cos(\theta)&\sin(\psi)\sin(\theta)\\
    0.0&-\sin(\theta)&\cos(\theta)
\end{bmatrix}.
\end{equation}

Since the transformation matrices for all pixels remain stationary, computational time is reduced by precomputing and storing the transformation matrices for all pixels beforehand.

As the detected surface can not face parallel or in the direction of the ray $(\vec{\mathbf{r}}=\mathbf{p}/|\mathbf{p}|)$, the normal vector is inversed when the dot product of the normal vector and the ray direction $(\mathbf{n}^c \centerdot \vec{\mathbf{r}})$ is positive. Finally, to verify whether the normal vector forms a consensus with neighboring points in the window, the distances between the points and the planes formed by their normal vectors and neighboring points are calculated. If the number of points with a point-to-plane distance within 5 cm is less than one-third of the window size, it is considered invalid. Through this process, the normal cloud $N$ containing the normal points $P$ is obtained. Figure~\ref{fig:normal_extraction} shows the overall process of normal extraction from the input cloud.

\subsection{Normal Cloud Registration}
\label{subsec:normal_cloud_registration}

The scan-to-scan matching may lead to unreliable registration result especially for the LiDAR motion in roll or pitch rotation and translation in scan rotation axis. Similar to LIO-SAM, we adopted the scan-to-submap matching method to address this issue. The submap is generated by accumulating the normal clouds of the preceding keyframes in the last keyframe coordinate system. For the last keyframe $\mathbb{K}_n$, the submap $\mathbf{N}_n$ augmenting previous $j$ keyframes is as follows:

\begin{equation} \label{eq3}
\mathbf{N}_{n} = N_n^n \cup N_{n-1}^n \cdots \cup N_{n-j}^n
\end{equation} 

where $N_i^k$ indicates normal cloud in keyframe $\mathbb{K}_i$ transformed into coordinate system of keyframe $\mathbb{K}_k$, and $\cup$ indicates the normal cloud augmentation.

For accurate correspondence search and fast matching, the current query frame $\mathbb{K}_q$ is transformed from its last obtained pose to the initial pose through IMU integration. Knowing the world coordinate system of both the target and query frames allows us to determine the initial relative pose between the two frames. Utilizing this information, the target frame is transformed into the coordinate system of the query frame and proceed with the matching process. Afterwards, for faster matching speed, the current normal cloud $N_{q}$ and the submap $\mathbf{N}_{n}$ are downsampled using a voxel grid filter.

To achieve stable matching between the resulting normal clouds, correspondences are established between pairs that satisfy the following two criteria: First, pairs for which the point-to-point distance falls within the distance threshold. Second, pairs for which the difference in direction of the normal vectors falls within the angle threshold. To accopmlish this, we first utilize a kd-tree to select submap points within the distance threshold for each query point in the current normal cloud. Then for the selected submap points in closest order, the angle difference in normal vector direction between the selected point and the query point is calculated. These two points are selected as correspondence pair if the angle difference falls within the angle threshold. 

Afterwards, the relative pose is calculated using the Gauss-Newton method by utilizing the correspondence pairs. The residual cost function for each pair is calculated as the point-to-plane distance, and the relative pose of the target frame from the query frame can be calculated by solving the following optimization problem:

\begin{equation}
    \label{eq:residual}
    (R, \mathbf{t}) = \underset{(R, \mathbf{t})}{\mathrm{argmin}} \sum_{k=0}^{m-1}(\mathbf{n}_{qk} \centerdot (R\mathbf{p}_{tk} + \mathbf{t} - \mathbf{p}_{qk}))^2.
\end{equation}

The resulting relative pose is then converted to relative pose factor and added to the factor graph.

\subsection{Loop Closure Detection}

Global loop detection algorithms often struggle in multifloor indoor environments characterized by repetitive structural features. This challenge is particularly evident in stairwells, where the recurring nature of their features can result in incorrect associations with clouds from different floors during inter-floor transitions. Therefore, in this study, we adopt a loop detection approach that focuses on matching keyframes within the vicinity of the current position rather than performing a global search. 

While techniques like ICP or GICP, which use radius search to find the closest points as correspondences, are frequently employed for local loop detection methods, they often lead to misalignment, especially in confined indoor spaces. This is mainly because indoor environments are typically composed of multiple segmented areas, leading to significant variations in LiDAR scans even with slight changes in LiDAR position. To address this issue, we introduce a viewpoint based loop closure detection method inspired by the projection techniques in \cite{SuMa} for better correspondence search.

\begin{figure}[t]
    \centering
    \begin{subfigure}[b]{1.0\linewidth}
        \centering
        \includegraphics[width=1.0\linewidth]{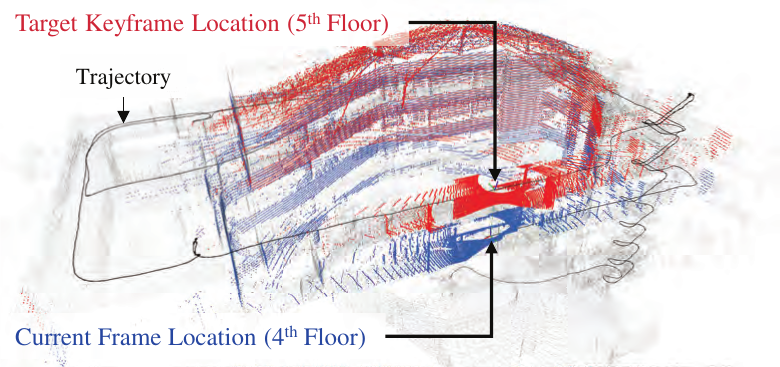}
        \caption{Target keyframe and current key frame on the map}
        \label{fig:inter_floor_loop_detection_ki}
    \end{subfigure}    

    \begin{subfigure}[b]{1.0\linewidth}
        \centering
        \includegraphics[width=1.0\linewidth]{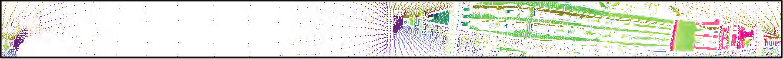}
        \caption{Downsampled current normal cloud projection}
        \label{fig:inter_floor_loop_detection_down_sample_curr_cloud}
    \end{subfigure}    

    \begin{subfigure}[b]{1.0\linewidth}
        \centering
        \includegraphics[width=1.0\linewidth]{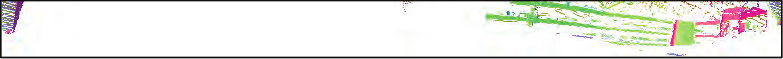}
        \caption{Target normal cloud projection in current frame}
        \label{fig:inter_floor_loop_detection_target_cloud_curr_frame}
    \end{subfigure}    

    \begin{subfigure}[b]{1.0\linewidth}
        \centering
        \includegraphics[width=1.0\linewidth]{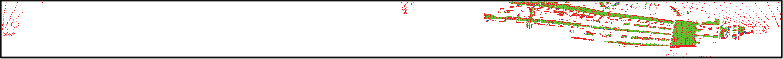}
        \caption{Matching results (green: accepted, red: rejected)}
        \label{fig:inter_floor_loop_detection_matching_result}
    \end{subfigure}    

    \caption{Inter-floor loop closure detection example using viewpoint based loop detection}
    \label{fig:inter_floor_loop_detection}
\end{figure}

Figure~\ref{fig:inter_floor_loop_detection} illustrates an example of viewpoint based loop detection. A kd-tree is constructed using the positions of each keyframe, and then the closest keyframe to the current frame is selected. In this process, keyframes immediately preceding the current frame are excluded from the kd-tree. Once a loop closure candidate keyframe is identified, the normal cloud of the candidate keyframe is transformed into the LiDAR pose of the current frame. Then, as shown in Fig~\ref{fig:inter_floor_loop_detection_target_cloud_curr_frame}, it is projected to match the current LiDAR viewpoint. For the points that are assigned to the same pixels during spherical projection, only the point with the closest range is assigned. 

This projection method, however, may not exclude points that should not be seen from the current viewpoint due to the sparse nature of the LiDAR point cloud. To address this issue, we utilize the normal points that lie within a 90-degree angle to the ray direction, denoted as $N^{+}$. As these points represent the opposite side of the wall, it is desirable to exclude any normal points that exceed a 90-degree angle to the ray direction, denoted as $N^{-}$, which pass through these points. Thus, for each pixel of the ${N^{+}}$ points, any nearby pixels of $N^{-}$ points are excluded if the $N^{-}$ point is further away than the $N^{+}$ point.

Subsequently, by filtering the projected points based on radial distance deviation and angular deviation compared to the downsampled current normal cloud projection image (Fig.~\ref{fig:inter_floor_loop_detection_down_sample_curr_cloud}), the correspondence pairs are obtained (Fig.~\ref{fig:inter_floor_loop_detection_matching_result}). Similar to the normal cloud registration method in \ref{subsec:normal_cloud_registration}, the matched point correspondence pairs are optimized to estimate the relative pose. These are then inserted into the pose-graph as loop closure factors.

\subsection{Degeneracy Detection}
\begin{figure}[t]
    \centering
    \includegraphics[width=1.0\linewidth]{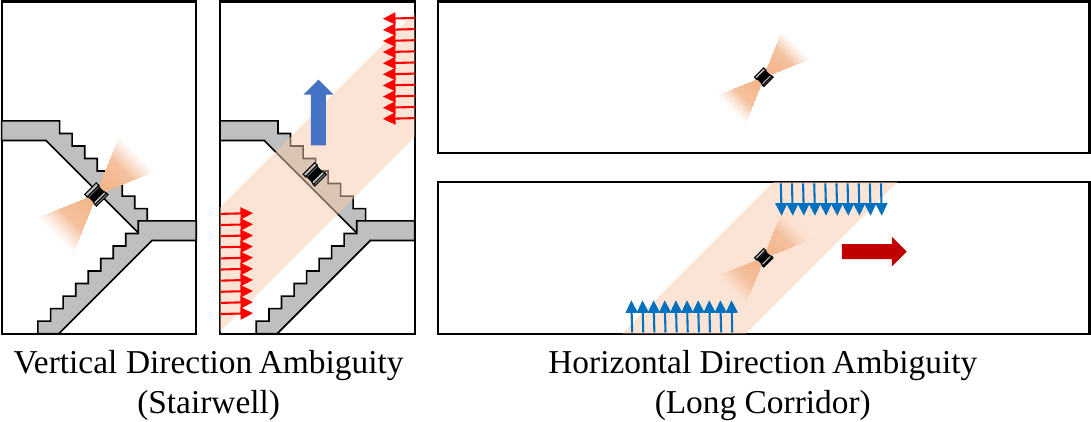}
    \caption{Degenerate cases in the stairwell and in the long corridor environment}
    \label{fig:degeneracy}
\end{figure}

In indoor environments, many surfaces are often arranged parallel to each other, leading to cases of degeneracy. Environments such as stairwells or long corridors, as illustrated in Fig.~\ref{fig:degeneracy}, exemplify such cases, where the normal vectors of surfaces are distributed in only two directions, resulting in translation ambiguity along the remaining direction. For instance, in the case of a stairwell, the normal vectors of the walls forming the stairwell are distributed horizontally, leading to high localization accuracy in the horizontal direction but potential ambiguity in the vertical direction.

To calculate a suitable matching uncertainty for such degeneracy situations, the distribution of normal vectors among the matched normal points can be utilized. This can be obtained through principal component analysis of the normal vectors as follows: First, the covariance matrix $C$ of the normal vectors are calculated as:

\begin{equation}
    \label{eq:covariance}
    \mathbf{C} = \left( \frac{1}{m} \sum_{k=0}^{m-1}(\mathbf{n}_{k}\mathbf{n}_{k}^{\top})\right).
\end{equation}

Subsequently, the covariance matrix $C$ is decomposed using eigenvalue decomposition as $C=V\Lambda V^{-1}$ where $V$ is the matrix composed of eigenvectors and $\Lambda$ is the matrix which the diagonal elements are eigenvalues as:

\begin{equation} \label{eigenvector}
\begin{split}
V & = \left[\mathbf{v}_{0}, \mathbf{v}_{1}, \mathbf{v}_{2}\right] \\
\Lambda & = diag(\lambda_0, \lambda_1, \lambda_2)
\end{split}
\end{equation}

where $\lambda_0 < \lambda_1 < \lambda_2$. The approximate distribution of normal vectors can be discerned using eigenvalues, wherein the smallest eigenvalue, $\lambda_0$, indicates a degenerate case if it falls below a specific threshold. Subsequently, each eigenvalue $\lambda_i$ corresponds to an eigenvector $\mathbf{v}_i$, allowing the measurement covariance $Q$ to be set as follows:

\begin{equation}
    \label{eq:eigenvector}
    Q = s ~ V diag\left(\frac{1}{\lambda_0}, \frac{1}{\lambda_1}, \frac{1}{\lambda_2} \right) V^{\top}
\end{equation}

where $s$ is a given constant. When matching with the immediately preceding keyframe, we inserted factors using this distribution-based measurement covariance in degenerate cases. However, in cases of loop closing where there is a high probability of incorrect matching, we refrained from inserting loop factors if degeneracy was detected to ensure stability.

\begin{figure}[tb]
    \centering
    \includegraphics[width=1.0\linewidth]{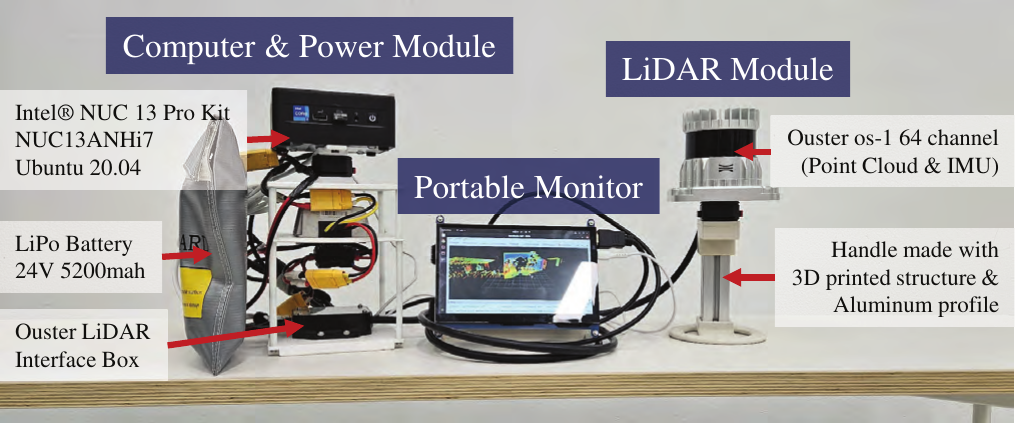}
    \caption{Data acquisition system for our own dataset}
    \label{fig:data_acquisition_system}
\end{figure}

\subsection{Pose-Graph SLAM Framework}
\label{subsection:poes_graph}
The pose-graph consists of prior factor given to the initial pose, relative pose factors obtained through the normal cloud registration, loop closure factors from the loop detection, and IMU factors and constant bias factors from the IMU-preintegration. The bias-reflected IMU measurements from the last pose-graph optimization result were integrated to continuously estimate the current frame at the IMU rate. If the pose difference between the current frame and the last frame exceeds a certain threshold, a new keyframe is inserted. The pose graph was constructed and optimized using the iSAM2 framework \cite{iSAM2}.

\subsection{Implementation Details}
The number of pixels used in spherical image projection was set to the channel number by 1024. For stable normal vector extraction, a 3 by 3 window was used for LiDAR with 32 channels or fewer, while a 5 by 5 window was used for LiDAR with more than 32 channels to compute the normal vectors. The distance threshold for normal cloud registration was set to 0.5 m, and the downsampling voxel size was set to 0.4 m or 0.2 m depending on the scenario. For keyframes, a new keyframe was added if the angle difference from the previous keyframe pose exceeded 30 degrees, or the distance difference is larger than a threshold. This distance threshold was set to be 1.0 m or 0.5 m depending on the building characteristics. The loop closure distance threshold was set to 10 meters.

\begin{figure}[tb]
    \centering
    \begin{subfigure}[b]{1.0\linewidth}
        \centering
        \includegraphics[width=1.0\linewidth]{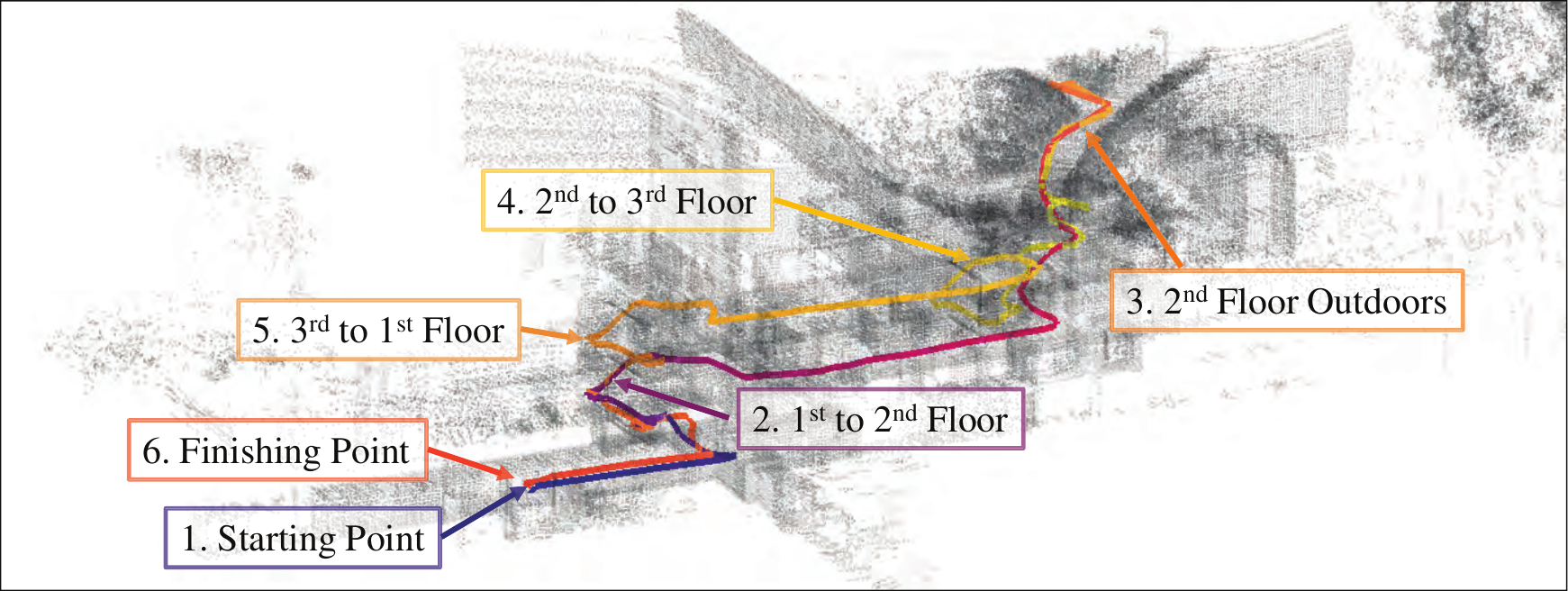}
        \caption{Data sequence}
        \label{fig:subt_result_data_sequence}
    \end{subfigure}    

    \begin{subfigure}[b]{1.0\linewidth}
        \centering
        \includegraphics[width=1.0\linewidth]{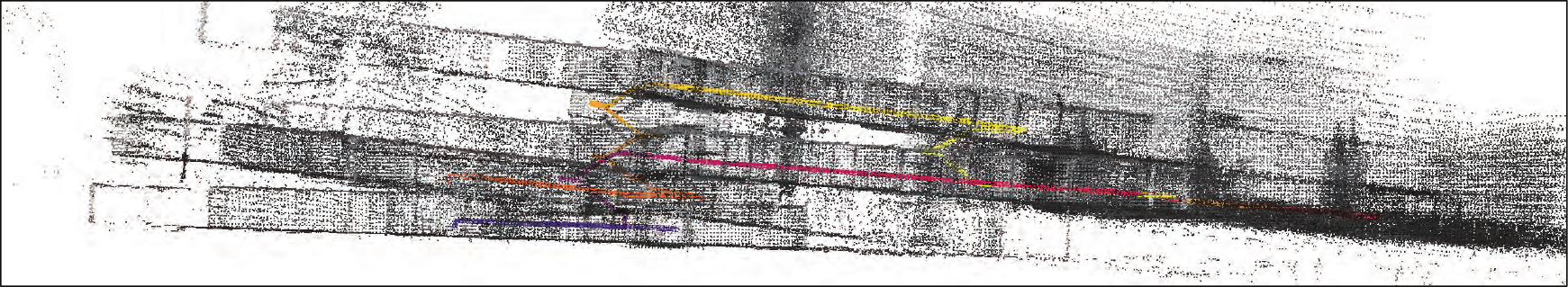}
        \caption{Faster-LIO}
        \label{fig:subt_result_faster_lio}
    \end{subfigure}    

    \begin{subfigure}[b]{1.0\linewidth}
        \centering
        \includegraphics[width=1.0\linewidth]{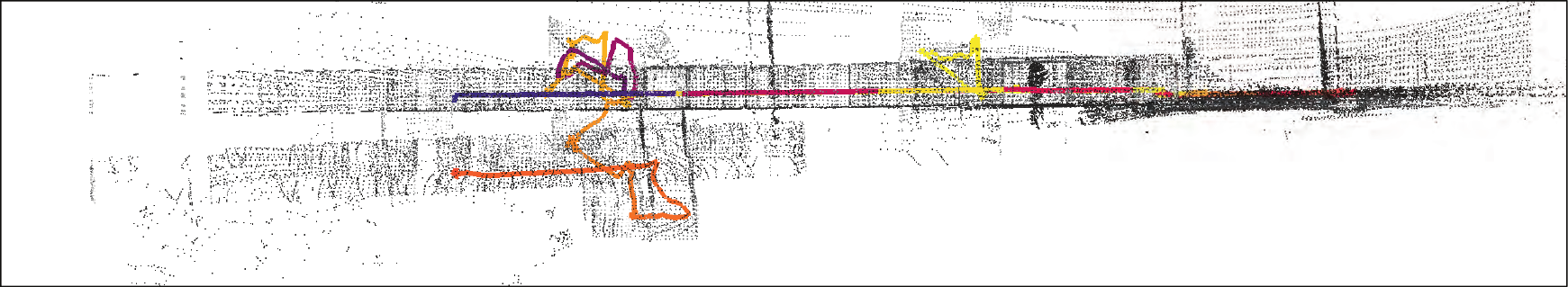}
        \caption{DLIO}
        \label{fig:subt_result_dlio}
    \end{subfigure}    

    \begin{subfigure}[b]{1.0\linewidth}
        \centering
        \includegraphics[width=1.0\linewidth]{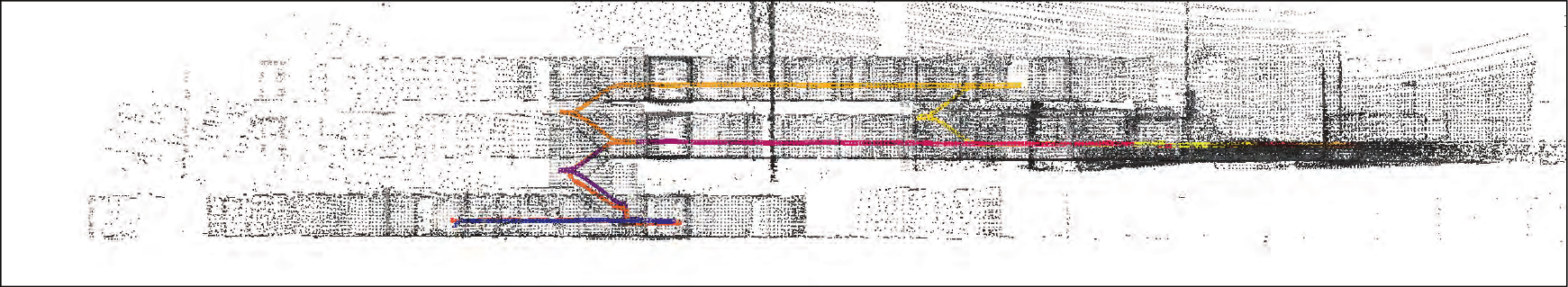}
        \caption{NV-LIO}
        \label{fig:subt_result_nv_lio}
    \end{subfigure}

    \caption{Sequence and results of the SubT-MRS Dataset}
    \label{fig:subt_result}
\end{figure}

\section{EXPERIMENTAL VALIDATION}
\label{sec:experimental_validataion}
To evaluate the performance of the proposed algorithm, we conducted tests using various datasets captured with different types of spinning LiDARs in diverse environments, including the SubT-MRS dataset \cite{SubT_MRS}, the Newer College dataset \cite{NewerCollege, NewerCollegeExtension}, and our own dataset. To evaluate the effectiveness of the proposed method in a multifloor indoor environment, the multifloor dataset of the SubT-MRS dataset was utilized for evaluation. This dataset comprises data obtained by a quadruped robot equipped with a 16-channel LiDAR and an IMU, traversing both the interior and exterior of a building from the first floor to the third floor. The Newer College dataset was captured in a campus environment using a handheld device, and LiDAR and its built-in IMU were used for evaluation. For our own dataset, the data was acquired in various buildings located in the KAIST campus using a handheld device as shown as Fig.~\ref{fig:data_acquisition_system}. Our algorithm was compared against state-of-the-art algorithms, including LIO-SAM \cite{LIO_SAM}, Fast-LIO2 \cite{FAST_LIO2}, Faster-LIO \cite{Faster_LIO}, and DLIO \cite{DLIO} which are publicly available. All tests were done on-line using a computer equipped with Intel i7-12700 CPU with 12 cores. 

\begin{table}[]
\captionof{table}{Result of Newer College Dataset \cite{NewerCollege}\label{table:NCD2020}}
\centering
\begin{tabular}{c||c|c|c|c|c}
\hline
\multirow{3}{*}{Algorithm} & \multicolumn{5}{c}{Root Mean Square Error (m)} \\ \cline{2-6} 
 & Short & Long & \begin{tabular}[c]{@{}c@{}}Dynamic\textbackslash\\ Spinning\end{tabular} & \begin{tabular}[c]{@{}c@{}}Quad w\textbackslash\\ Dynamics\end{tabular} & \begin{tabular}[c]{@{}c@{}}Parkland\\ Mound\end{tabular} \\ \hline
LIO-SAM & 0.4052 & 0.5645 & 0.1204 & \textbf{0.1219} & 0.1523 \\ \hline
Fast-LIO2 & 0.4236 & \textbf{0.3781} & 0.1335 & 0.2944 & \textbf{0.1329} \\ \hline
Faster-LIO & 0.5243 & 1.6950 & 0.1240 & 0.2047 & 0.2164 \\ \hline
DLIO & \textbf{0.3834} & 0.4065 & 0.1350 & 0.1557 & 0.1407 \\ \hline
\textbf{NV-LIO} & 0.4795 & 0.4632 & \textbf{0.1152} & 0.1412 & 0.2799 \\ \hline
\end{tabular}
\end{table}

\begin{table*}[tbh]
\captionof{table}{Result of Newer College Dataset Extension \cite{NewerCollegeExtension}\label{table:NCD2021}}
\centering
\begin{tabular}{c||c|c|c|c|c|c|c|c|c|c|c|c|c}
\hline
\multirow{3}{*}{Algorithm} & \multicolumn{12}{c|}{Root Mean Square Error (m)} & \multirow{3}{*}{\begin{tabular}[c]{@{}c@{}}Avg.\\ Comp.\\(ms) \end{tabular}}\\ \cline{2-13} 
 & \begin{tabular}[c]{@{}c@{}}Quad\\ Easy\end{tabular} & \begin{tabular}[c]{@{}c@{}}Quad\\ Medium\end{tabular} & \begin{tabular}[c]{@{}c@{}}Quad\\ Hard\end{tabular} & Stairs & Cloister & Park & \begin{tabular}[c]{@{}c@{}}Math\\ Easy\end{tabular} & \begin{tabular}[c]{@{}c@{}}Math\\ Medium\end{tabular} & \begin{tabular}[c]{@{}c@{}}Math\\ Hard\end{tabular} & \begin{tabular}[c]{@{}c@{}}Under-\\ Easy\end{tabular} & \begin{tabular}[c]{@{}c@{}}Under-\\ Medium\end{tabular} & \begin{tabular}[c]{@{}c@{}}Under-\\ Hard\end{tabular} & \\ \hline
LIO-SAM & 0.0743 & \textbf{0.0668} & \textbf{0.1260} & failed & \textbf{0.0741} & 0.3683 & 0.0819 & 0.1259 & \textbf{0.0852} & \textbf{0.0578} & \textbf{0.0645} & 0.4188 & 113.7 \\ \hline
Fast-LIO2 & 0.0830 & 0.0823 & 0.1843 & 0.1130 & 0.1229 & 0.3527 & 0.1015 & 0.1249 & 0.1336 & 0.1066 & 0.1050 & 0.1040 & \textbf{22.74} \\ \hline
Faster-LIO & 0.2631 & 0.1624 & 0.1756 & 0.1171 & 0.1711 & 0.4086 & 0.1740 & 0.1899 & 0.0937 & 0.0597 & 0.0805 & 0.0849 & 22.92 \\ \hline
DLIO & \textbf{0.0734} & 0.0691 & 0.1655 & 0.1531 & 0.0935 & 0.3185 & 0.1141 & 0.1066 & 0.0853 & 0.0642 & 0.0655 & 0.0828 & 34.52 \\ \hline
\textbf{NV-LIO} & 0.0762 & 0.0757 & 0.1780 & \textbf{0.0876} & 0.0767 & \textbf{0.2896} & \textbf{0.0616} & \textbf{0.0952} & 0.0936 & 0.0622 & 0.0718 & \textbf{0.0780} & 54.25 \\ \hline
\end{tabular}
\end{table*}

\begin{figure*}[tbh]
    \centering
    \begin{subfigure}[b]{0.19\linewidth}
        \centering
        \includegraphics[width=1.0\linewidth]{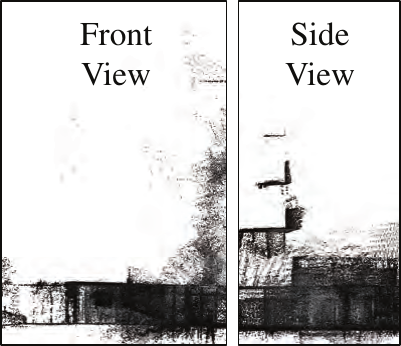}
        \caption{LIO-SAM}
        \label{fig:ki_stairwell_lio_sam}
    \end{subfigure}
    \hfill
    \begin{subfigure}[b]{0.19\linewidth}
        \centering
        \includegraphics[width=1.0\linewidth]{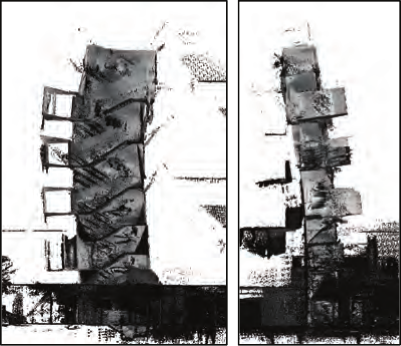}
        \caption{Fast-LIO2}
        \label{fig:ki_stairwell_fast_lio2}
    \end{subfigure}
    \hfill
    \begin{subfigure}[b]{0.19\linewidth}
        \centering
        \includegraphics[width=1.0\linewidth]{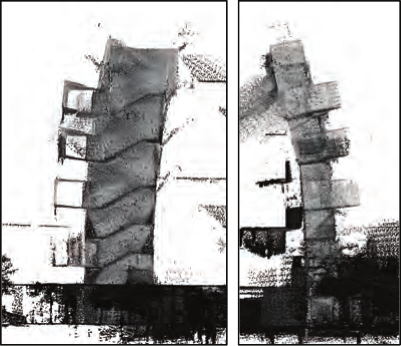}
        \caption{Faster-LIO}
        \label{fig:ki_stairwell_faster_lio}
    \end{subfigure}
    \hfill
    \begin{subfigure}[b]{0.19\linewidth}
        \centering
        \includegraphics[width=1.0\linewidth]{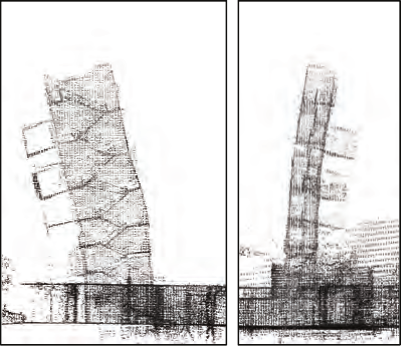}
        \caption{DLIO}
        \label{fig:ki_stairwell_dlio}
    \end{subfigure}
    \hfill
    \begin{subfigure}[b]{0.19\linewidth}
        \centering
        \includegraphics[width=1.0\linewidth]{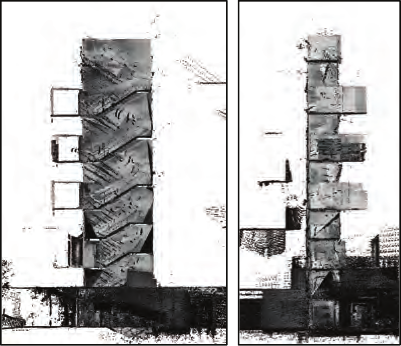}
        \caption{NV-LIO}
        \label{fig:ki_stairwell_nv_lio}
    \end{subfigure}
    
    \begin{subfigure}[b]{0.49\linewidth}
        \centering
        \includegraphics[width=1.0\linewidth]{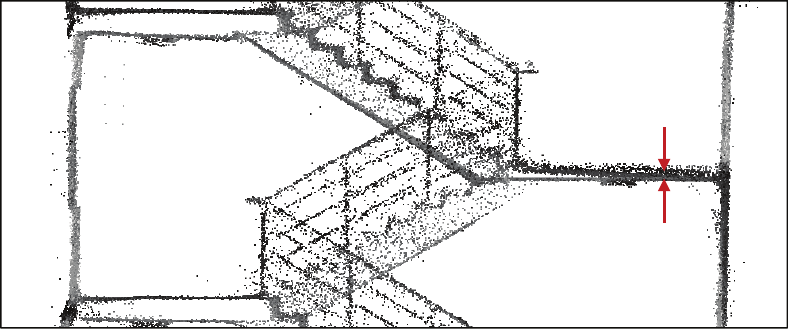}
        \caption{Cross-section of Faster-LIO mapping result}
        \label{fig:ki_stairwell_faster_lio_cross}
    \end{subfigure}    
    \hfill
    \begin{subfigure}[b]{0.49\linewidth}
        \centering
        \includegraphics[width=1.0\linewidth]{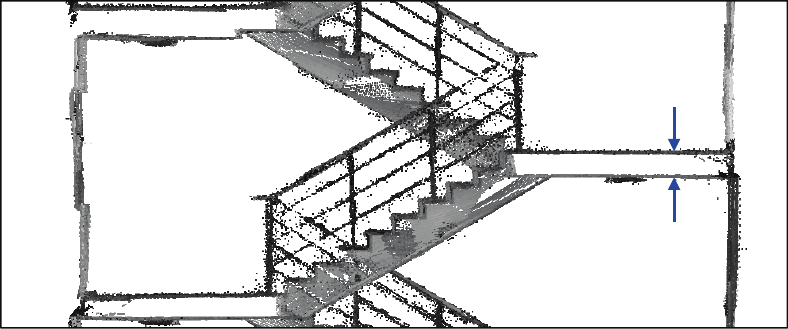}
        \caption{Cross-section of NV-LIO mapping result}
        \label{fig:ki_stairwell_nv_lio_cross}
    \end{subfigure}    

    \caption{Mapping results of the own dataset (stairwell)}
    \label{fig:ki_stairwell}
\end{figure*}

\begin{figure*}[tbh]
    \centering
    \begin{subfigure}[b]{0.325\linewidth}
        \centering
        \includegraphics[width=1.0\linewidth]{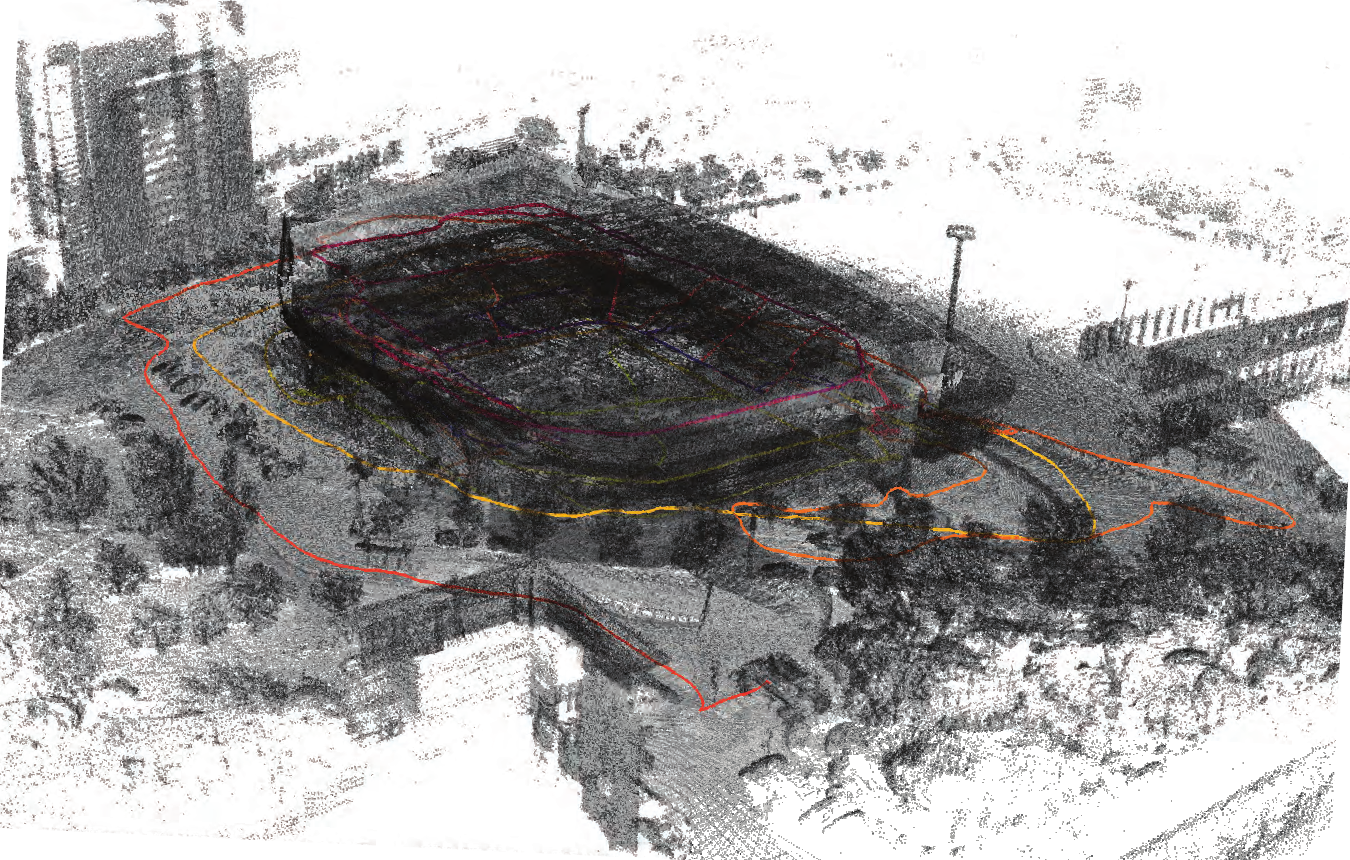}
        \caption{Sports Complex Building}
        \label{fig:kaist_mapping_sc}
    \end{subfigure}    
    \hfill
    \begin{subfigure}[b]{0.325\linewidth}
        \centering
        \includegraphics[width=1.0\linewidth]{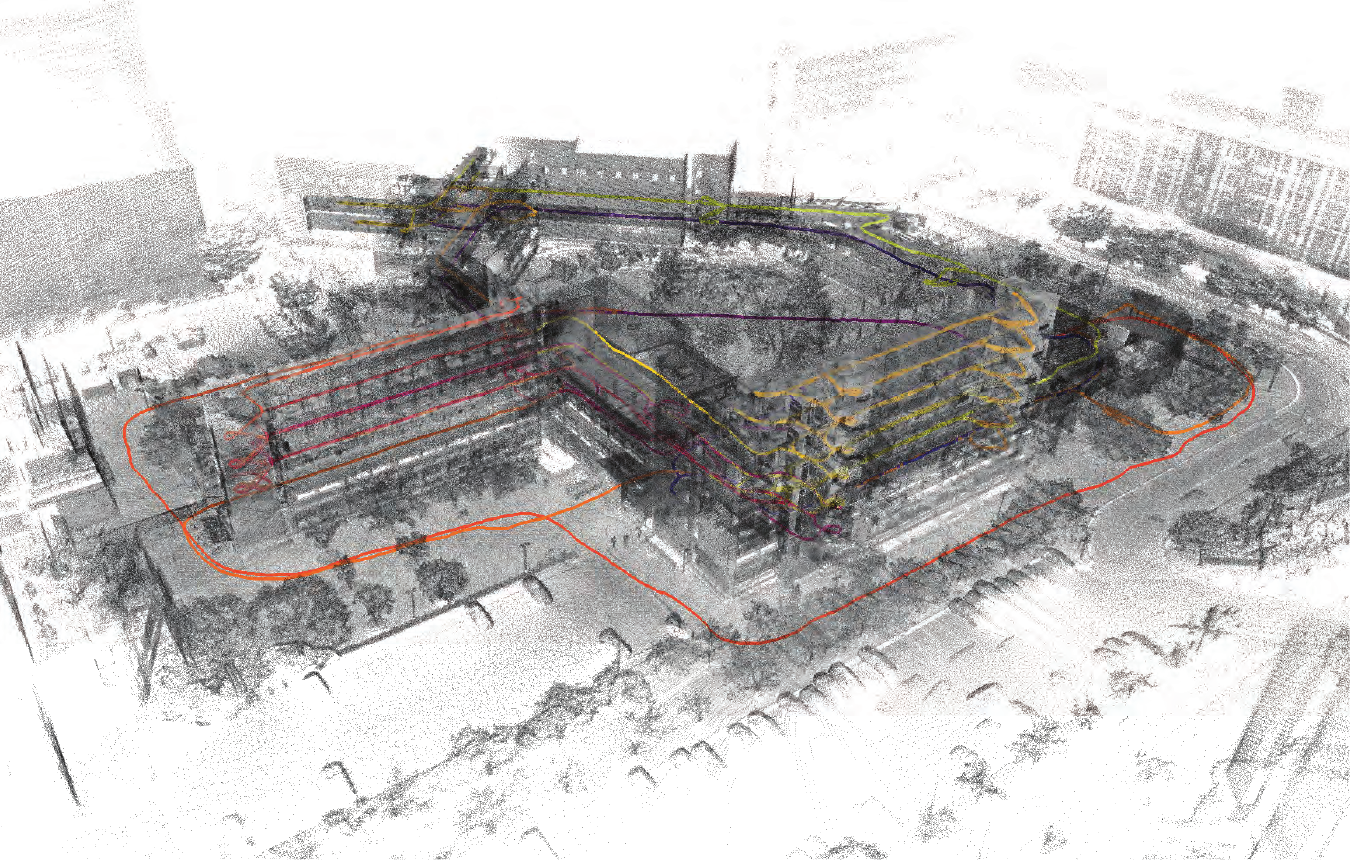}
        \caption{E3 Building}
        \label{fig:kaist_mapping_e3}
    \end{subfigure}    
    \hfill
    \begin{subfigure}[b]{0.325\linewidth}
        \centering
        \includegraphics[width=1.0\linewidth]{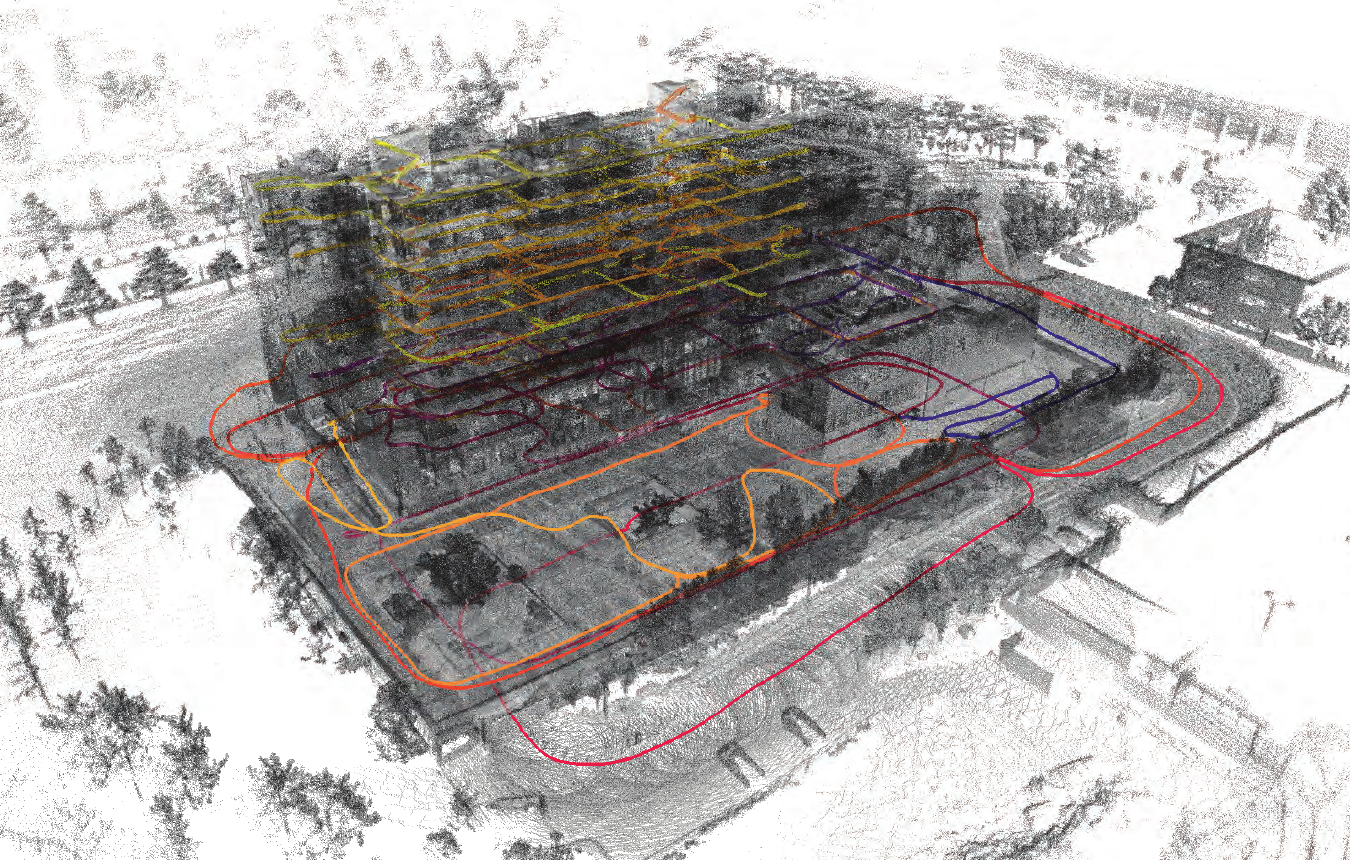}
        \caption{N1 Building}
        \label{fig:kaist_mapping_n1}
    \end{subfigure}    
    \caption{Mapping result of buildings with different characteristics}
    \label{fig:kaist_mapping}
\end{figure*}

\subsection{Results}

\subsubsection{SubT-MRS Multifloor Dataset}
Due to the absence of ground-truth trajectory in the SubT-MRS dataset, qualitative results are presented as shown in Figure~\ref{fig:subt_result}. Considering narrow corridors and staircases, the voxel size for all algorithms was set to 0.2 m, and the keyframe distance for keyframe-based algorithms (LIO-SAM, DLIO, and NV-LIO) was set to 0.5 m. During evaluation, Fast-LIO2 and LIO-SAM failed and were not included in the results. Faster-LIO exhibited significant drift each time it passed the stairs, resulting in incorrect map formation, as shown in Fig.~\ref{fig:subt_result_faster_lio}. DLIO produced incorrect mapping as observed in the trajectory on the stairs, where keyframes from different floors were incorrectly matched (Fig.~\ref{fig:subt_result_dlio}). In contrast, as shown in Fig.~\ref{fig:subt_result_nv_lio}, NV-LIO clearly delineated the boundaries of each floor, providing good state estimation even in staircase environments, resulting in overall well-formed mapping result.

\subsubsection{Newer College Dataset}
The results for the Newer College dataset are shown in Table~\ref{table:NCD2020} and Table~\ref{table:NCD2021}. The downsampling voxel size for all algorithms was fixed at 0.4 m, and the keyframe insertion distance for NV-LIO was set at 1.0 m for evaluation. The evaluations were conducted with the provided ground truth trajectory aligned to the IMU frame. The absolute pose error was measured using the translation root mean square error after aligning the trajectory with Umeyama \cite{Umeyama} alignment using evo \cite{EVO}. 

The results indicate that NV-LIO requires additional pre-processing, leading to increased average computational time. However, it demonstrates similar or lower errors compared to state-of-the-art methods in outdoor environments, while showing greater robustness in indoor settings. Particularly, it performs well in very narrow environments such as the stairs dataset or in scenarios with rapid motion, as observed in the underground-hard dataset.

\subsubsection{Dataset of KAIST Buildings}

Figure~\ref{fig:ki_stairwell} illustrates the results for the stairwell dataset of the 5-floor building, which was shown in Figure~\ref{fig:ki_building}. LIO-SAM failed to register as soon as it entered the stairwell, resulting in mapping failure to the extent that further progress was impossible (Fig.\ref{fig:ki_stairwell_lio_sam}). While Fast-LIO2 and Faster-LIO did not fail, but the errors accumulated gradually as ascending the stairs, resulting curved mapping result of the stairwell (Fig.~\ref{fig:ki_stairwell_fast_lio2} and \ref{fig:ki_stairwell_faster_lio}). DLIO suffered wrong keyframe matching at the beginning, resulting one floor missing from the whole stairwell. It also had the same problem of curved mapping due to accumulation of the errors (Fig.~\ref{fig:ki_stairwell_dlio}). On the other hand, the mapping result using NV-LIO (Fig.~\ref{fig:ki_stairwell_nv_lio}) shows vertical structure, preserving the actual shape of the stairwell. Figure~\ref{fig:ki_stairwell_faster_lio_cross} and \ref{fig:ki_stairwell_nv_lio_cross} represent magnified views of the results obtained from Faster-LIO and NV-LIO, respectively. As indicated by the arrows, while Faster-LIO failed to map properly due to incorrect correspondence between the ceiling of the lower floor and the floor of the upper floor, NV-LIO considers the directions of the normal vectors, avoiding mismatches and leading to correct results.

Additionally, NV-LIO was assessed in buildings with varying characteristics. Figure~\ref{fig:kaist_mapping_sc} depicts a stadium-shaped building with an underground parking lot and a three story structure with an open center. Figure~\ref{fig:kaist_mapping_e3} shows a building comprising research labs and classrooms across five and six stories, connected to the three-story building. Figure~\ref{fig:kaist_mapping_n1} represents a research facility consisting of an underground parking lot and nine above-ground stories. NV-LIO successfully conducted online SLAM for each dataset, lasting approximately one hour, allowing for qualitative evaluation of the mapping results. All dataset mapping sequences and results are available in our code repository.

\section{CONCLUSIONS}
\label{sec:conclusion}

This paper introduces NV-LIO, a normal vector based tightly-coupled LiDAR-inertial odometry framework designed for indoor SLAM applications. NV-LIO utilizes the normal vectors extracted from the LiDAR scans for cloud registration, degeneracy detection and loop closure detection to ensure robust SLAM performance in narrow indoor environments. The proposed method was evaluated through public datasets and our own datasets encompassing various types of buildings. The experimental results demonstrate that NV-LIO outperforms existing methods in terms of accuracy and robustness, particularly in challenging indoor scenarios such as narrow corridors and staircases.

\addtolength{\textheight}{-12cm}   % This command serves to balance the column lengths

%%%%%%%%%%%%%%%%%%%%%%%%%%%%%%%%%%%%%%%%%%%%%%%%%%%%%%%%%%%%%%%%%%%%%%%%%%%%%%%%

%%%%%%%%%%%%%%%%%%%%%%%%%%%%%%%%%%%%%%%%%%%%%%%%%%%%%%%%%%%%%%%%%%%%%%%%%%%%%%%%

%%%%%%%%%%%%%%%%%%%%%%%%%%%%%%%%%%%%%%%%%%%%%%%%%%%%%%%%%%%%%%%%%%%%%%%%%%%%%%%%

% \section*{ACKNOWLEDGMENT}

% \bibliographystyle{ieeetr}
% \bibliography{bibfile}

\end{document}